\documentclass[runningheads]{llncs}
\usepackage[T1]{fontenc}
\usepackage{graphicx}
\usepackage{amsfonts}
\usepackage[strings]{underscore}
\usepackage{array}
\usepackage{makecell}
\usepackage{caption}
\usepackage{subcaption}
\usepackage[export]{adjustbox}
\usepackage{interval}
\usepackage{amsmath,amssymb}
\usepackage{multirow}
\usepackage{hyperref}
\usepackage[numbers]{natbib}

\begin{document}
\title{Neural Field Conditioning Strategies for~2D~Semantic~Segmentation\thanks{The authors gratefully acknowledge support from the
DFG (CML, MoReSpace, LeCAREbot), BMWK (SIDIMO, VERIKAS), and the European
Commission (TRAIL, TERAIS).}}
\titlerunning{Neural Fields for 2D Semantic Segmentation}
%
\author{Martin Gromniak\inst{1,2} \and
Sven Magg\inst{3} \and
Stefan Wermter\inst{1}}
\authorrunning{M. Gromniak et al.}
%
\institute{Knowledge Technology, Department of Informatics, University of Hamburg 
\and 
ZAL Center of Applied Aeronautical Research
\and
Hamburger Informatik Technologie-Center e.V. (HITeC)
}

\maketitle              
\begin{abstract}
Neural fields are neural networks which map coordinates to a desired signal. When a neural field should jointly model multiple signals, and not memorize only one, it needs to be conditioned on a latent code which describes the signal at hand. Despite being an important aspect, there has been little research on conditioning strategies for neural fields. In this work, we explore the use of neural fields as decoders for 2D semantic segmentation. For this task, we compare three conditioning methods, simple concatenation of the latent code, Feature Wise Linear Modulation (FiLM), and Cross-Attention, in conjunction with latent codes which either describe the full image or only a local region of the image. Our results show a considerable difference in performance between the examined conditioning strategies. Furthermore, we show that conditioning via Cross-Attention achieves the best results and is competitive with a CNN-based decoder for semantic segmentation.

\end{abstract}
\section{Introduction}

Lately, neural networks for semantic segmentation have been mostly based on the fully convolutional network (FCN) \cite{longFullyConvolutionalNetworks2015} paradigm. FCN models typically consist of an encoder and a decoder which are both build with stacked convolution layers. The purpose of the encoder is to extract features from the image. With increasing depth of the encoder, the features get more abstract and the resolution of the feature maps is progressively reduced. The decoder on the other hand takes the low-resolution feature map from the encoder as an input and upscales them to the resolution of the original image so that pixel-level classification can be performed. 

While encoders in the form of convolutional neural networks (CNN) have been rigorously studied, considerably less research has been conducted on the decoder side of semantic segmentation networks. The main challenge for the decoder is to upscale the feature map to the image's original resolution and simultaneously produce accurate region borders. In CNN-based decoders, typically upsampling or transposed convolution operators are used to progressively increase the spatial resolution of the feature maps. These operations introduce a particular kind of inductive bias. For example, transposed convolutions can create spectral artifacts in the upscaled feature maps \cite{durallWatchYourUpConvolution2020}. Another apparent disadvantage of CNN decoders is that they struggle to capture long-range dependencies between different parts of the image, due to their locally connected structure. 

In the last few years, neural fields, aka implicit neural representations or coordinate-based networks, have received much attention for learning a variety of different signals, for example, 1D audio signals \cite{sitzmannImplicitNeuralRepresentations2020}, 2D images \cite{chenLearningContinuousImage2021,tancikFourierFeaturesLet2020} and 3D geometries \cite{meschederOccupancyNetworksLearning2019,sitzmannSceneRepresentationNetworks2020}. A neural fields takes (spatial) coordinates $x \in \mathbb{R}^d$ as input and maps them to a task-dependent signal $y = \Phi(x)$ through a neural network. For example, a neural field representing an RGB image takes 2D image coordinates as input and produces three RGB values at each location.  One interesting property of neural fields is that they represent signals as continuous functions on their respective spatial domain. 

Inspired by the recent successes of neural fields, we explore the use of neural fields as decoders in semantic segmentation networks. In this regard, we hypothesize that (continuous) neural fields provide an inductive bias which can be better suited for reconstructing high-resolution semantic maps compared with (discrete) CNN-based decoders. In our work, we examine multiple conditioning strategies, which enable the neural field decoder to make use of the information in the latent feature map produced by the encoder. Through our comparative study, we aim to provide more insights into conditioning methods of neural fields, as research have been extremely sparse in this regard. Furthermore, we believe that 2D semantic segmentation provides a well-defined task for studying conditioning methods, as it has comprehensive metrics and the possibility for insightful visualizations of the learned geometries.

\section{Related Work}

\textbf{Semantic Segmentation} Encoder-decoder fully convolutional networks  \cite{longFullyConvolutionalNetworks2015} have become the predominant approach for semantic segmentation. They share the challenge how to encode high-level features in typically low-resolution feature maps and subsequently upscale these feature maps to retrieve pixel-accurate semantic predictions. Multiple approaches \cite{ronnebergerUNetConvolutionalNetworks2015,badrinarayananSegNetDeepConvolutional2016} have introduced skip-connections between the encoder and the decoder, which help the decoder to combine global with local information and therefore recover sharp object boundaries.   
One drawback of CNNs is that, because of their locally connected structure, they struggle to combine information which is spatially distributed across the feature maps. Research attempting to mitigate this drawback has proposed attention mechanisms over feature maps to selectively capture and combine information on a global scale \cite{fuDualAttentionNetwork}. Extending the concept of attention further, neural network architectures based fully on transformers have been proposed recently for semantic segmentation \cite{strudelSegmenterTransformerSemantic2021}. In our work, we utilize a CNN, which is more efficient than transformers, for extracting features and use attention in one of our conditioning methods.

\bigbreak
\noindent\textbf{CNN Decoders} Research on decoders has been more sparse than research on neural network encoders, i.e. CNN backbones. Wojna et al. \cite{wojnaDevilDecoder2017a} compared different CNN-based decoders for several pixel-wise prediction tasks and observed significant variance in results between different types of decoders. Multiple works \cite{odena2016deconvolution,durallWatchYourUpConvolution2020} have provided evidence that upscaling using transposed convolution operators introduce artifacts in the feature maps and therefore the decoder's output. We aim to avoid any explicit or implicit discrete artifacts by using a continuous neural field decoder. 

\bigbreak
\noindent\textbf{Neural Fields} Neural fields were introduced in 2019 as a representation for learning 3D shapes \cite{meschederOccupancyNetworksLearning2019,parkDeepSDFLearningContinuous2019}. Following works extended neural fields by learning colored appearance of scenes and objects \cite{sitzmannSceneRepresentationNetworks2020,mildenhallNeRFRepresentingScenes2020}. Particularly NeRF \cite{mildenhallNeRFRepresentingScenes2020} has attracted a lot of attention, as it is able to generate very realistic novel views of a scene, learning from images and associated poses. NeRF effectively overfits a neural network for one individual scene. This limits the usability as the neural field needs to be re-trained for every new scene. Some works have explored the use of neural fields for semantic segmentation. Vora et al. \cite{voraNeSFNeuralSemantic2021} built a 3D segmentation on top of the NeRF approach. Hu et al. \cite{huLearningImplicitFeature2022} used neural fields in conjunction with CNNs for upsampling and aligning feature maps in the decoder of a semantic segmentation network.  

\bigbreak
\noindent\textbf{Neural Field Conditioning} When a neural field should share knowledge between different signals, it needs to be conditioned on a latent code which describes the signal at hand. Several conditioning approaches have been explored in the literature. Methods based on global conditional codes use one code to describe the whole signal \cite{meschederOccupancyNetworksLearning2019,sitzmannLightFieldNetworks2022}. Methods based on local conditional codes \cite{yuPixelNeRFNeuralRadiance2021,chenLearningContinuousImage2021} use a different code for each spatial area in the signal. On top of these, there exist multiple methods how a neural field can actually consume a conditional code, which we describe in detail in Section 3.3.  Rebain et al. \cite{rebainAttentionBeatsConcatenation2022} compared different methods for conditioning neural fields for 2D and 3D tasks, but did not consider global and local conditional codes. In the neural field community, there is a lack of comparative research on what conditioning strategies work well for which task. We attempt to shed more light on this by comparing different conditioning strategies for the well-defined task of 2D semantic segmentation.

\section{Method}

\begin{figure}[t]
\includegraphics[width=\textwidth]{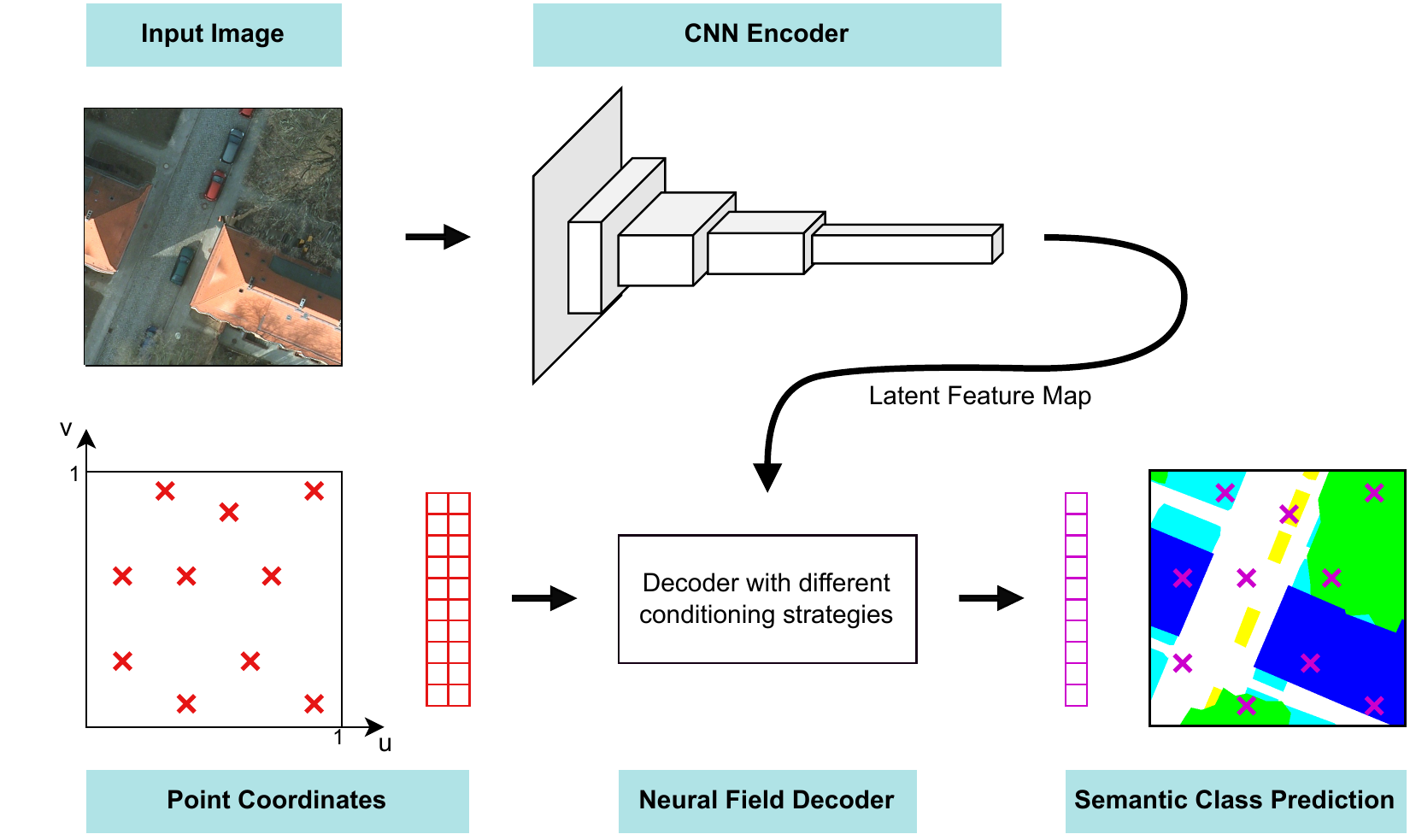}
\caption{Our high level neural network architecture. A CNN encoder encodes an image into a feature map. During training, $S$ points per image are sampled within the image (red) and fed into the decoder. The decoder is a conditional neural field for which we use different conditioning strategies. For every point the decoder outputs a prediction of the semantic class at this position (purple).} \label{fig1}
\end{figure}

\subsection{Neural Network Architecture and Training Procedure}

Our high-level architecture involves a CNN decoder and a neural field decoder(see Figure \ref{fig1}). We use a CNN to efficiently encode an image into a feature map with size $c \times h \times w$, where $c$ is the number of channels, $w$ is the spatial width and $h$ is the spatial height. From this feature map, we calculate the conditional code for the neural field decoder in different ways, depending on the conditioning strategy. During training, for every image, we sample $S$ random points within the image. At test time, the points are densely sampled so that there exists one point for each pixel. The point coordinates are normalized to the $\interval{0}{1}$ range, stacked, and fed to the neural field decoder as input. For every point, the decoder predicts the semantic class at that position in the image. We use a cross-entropy loss to train the whole setup in an end-to-end fashion. At test time, the class predictions per point are arranged into an image. Thereby, we can compare the predicted feature map with the labeled feature map using standard image segmentation metrics.

\subsection{Latent Code Source: Global vs. Local}

\begin{figure}
\includegraphics[width=\textwidth]{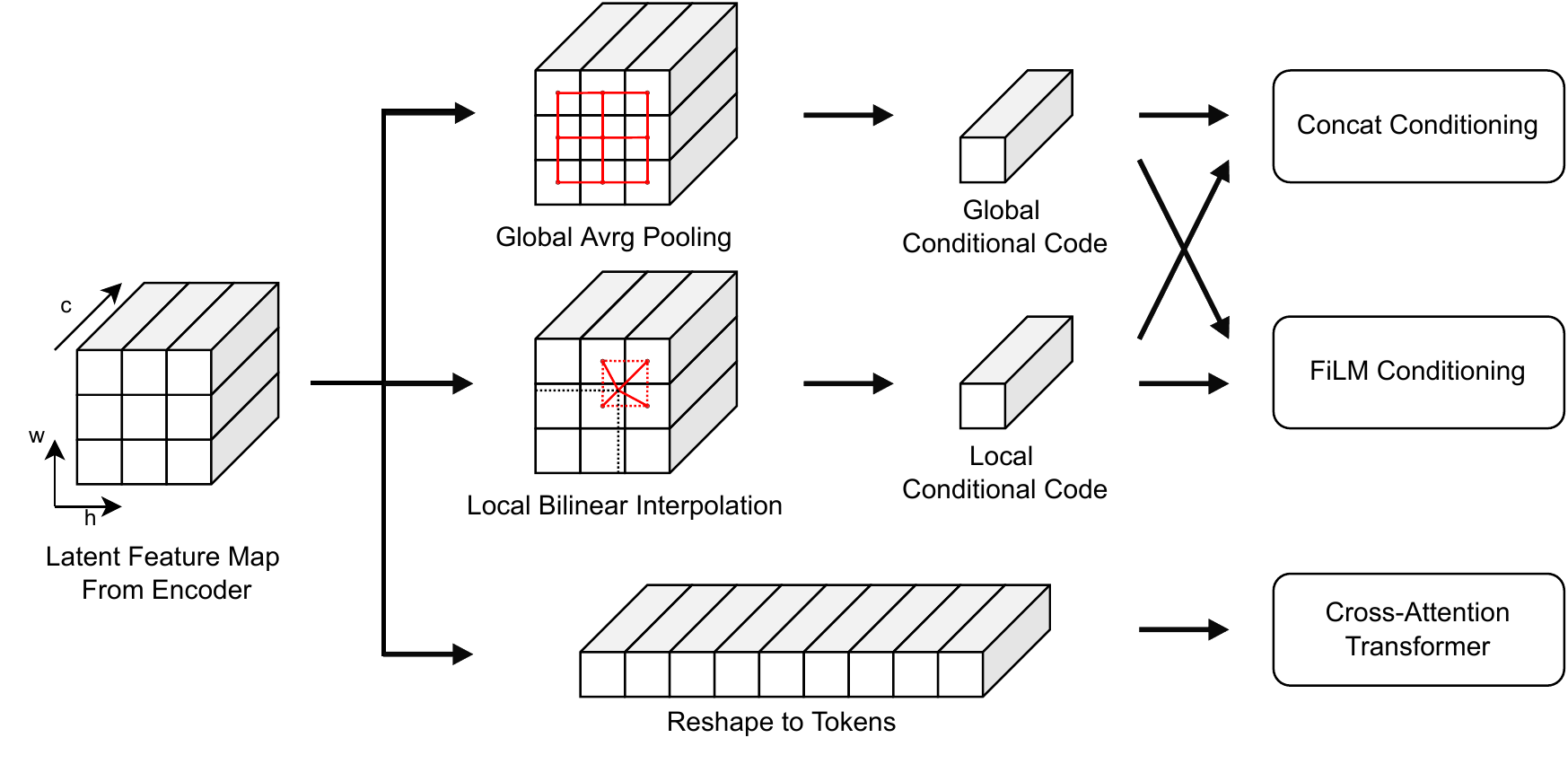}
\caption{A visualization of our conditioning strategies. We consider three conditioning methods: Concat conditioning, FiLM conditioning and Cross-Attention conditioning (right side). For Concat and FiLM conditioning, one feature vector is used, which can be calculated from global features (top path) or local features (mid path). The input to the Cross-Attention Transformer is the whole feature map, which is reshaped and treated as tokens (bottom path).} \label{fig2}
\end{figure}

First, we differentiate how the conditional code is calculated based on the feature map from the encoder. We consider a \textit{global} code and a \textit{local} code. The global code represents the content of the complete image. Naturally, it can capture the global context in the image well. However, due to its limited capacity, it might not be able to capture fine, local geometries. On the other hand, the local code represents a spatially limited area in the image. It can utilize its full capacity for modeling the geometry in one area with high fidelity, however, it might lack global context. For example, the probability of detecting a car rises when a street is detected somewhere in the image.   
 
We calculate the global code vector by applying a global average pooling operation. It averages all the entries in the feature map across the spatial dimensions (see the top path in Figure 2). This is a standard operation which is used, for example, in the ResNet classification head \cite{heDeepResidualLearning2016}. Through this procedure, we calculate one global code per image.  

For calculating the local code, we utilize the point coordinates, in addition to the feature map. For every point, we "look up" the value of the feature map at this position. For this purpose, we normalize the feature map's spatial dimensions to the [0,1] range, and therefore effectively align it with the input image. We then perform a bilinear interpolation within the feature maps based on the point coordinate to calculate the local code vector (see the middle path in Figure 2). As a result, we have $S$ local codes per image, one for every point.

In addition to using either a global or a local code, we also consider the combination of both to jointly exploit their individual advantages. We do this by concatenating both codes.

\subsection{Conditioning the Neural Field Decoder}

Conditioning a neural field enables it to effectively adapt the knowledge which is shared across all signals to the signal at hand.

\subsubsection{Conditioning by Concatenation}

In the simplest conditioning method, the conditional code is concatenated to the point coordinates and is jointly used as input to the neural field. We re-concatenate the conditional code to other hidden layers using skip connections. This approach is used by HyperNeRF \cite{parkHyperNeRFHigherDimensionalRepresentation2021}. It has the advantage of being conceptually simple, however, it is computationally inefficient \cite{rebainAttentionBeatsConcatenation2022}, because it requires $O(k(c+k))$ parameters for the fully connected layers in the neural field, where $k$ is the hidden layer width and $c$ is the size of the conditioning vector.

\subsubsection{Feature-Wise Linear Modulation}

Another way to condition a neural field is to use the latent code together with an MLP to regress the parameters of the neural field. When all parameters of the neural field are calculated in this way, the approach is known as hyper-networks \cite{haHyperNetworks2016}. Feature-wise Linear Modulation (FiLM) \cite{perezFiLMVisualReasoning2018} is a more constraint subtype of hyper-networks where, instead of predicting all the weights, feature-wise modulations of activations in the neural field are predicted. This approach is used in Occupancy Networks \cite{meschederOccupancyNetworksLearning2019} and piGAN \cite{chanPiGANPeriodicImplicit2021}.

\subsubsection{Cross-Attention}

Conditioning by Cross-Attention has been introduced by Jiang et al. \cite{jiangCOTRCorrespondenceTransformer2021} and was extended in the Scene Representation Transformer \cite{sajjadiSceneRepresentationTransformer2022}. The core idea is to selectively attend to features at different spatial positions, based on the point coordinates. A transformer architecture with Cross-Attention layers is used where the queries are derived from the point coordinates and the feature maps serve as a set of tokens. This approach does have an interesting connection with using local codes, as both approaches calculate a feature vector by weighting entries in the feature maps based on the current point coordinate. However, in difference to the spatial "look up" of local codes, which can be performed for free, the Cross-Attention operation can flexibly query both local and global information as needed at the cost of more computation \cite{rebainAttentionBeatsConcatenation2022}.

\section{Experiments}

We evaluate seven conditioning strategies on a publicly available dataset for semantic segmentation. Concat conditioning and FiLM conditioning are used in conjunction with global, local and combined conditional codes each. The Cross-Attention Transformer uses the reshaped feature map as input (see Figure \ref{fig2}).

\subsection{Dataset}

For our experiments, we used the Potsdam dataset\footnote{\href{https://www.isprs.org/education/benchmarks/UrbanSemLab/2d-sem-label-potsdam.aspx}{https://www.isprs.org/education/benchmarks/UrbanSemLab/2d-sem-label-potsdam.aspx}}  which is part of the ISPRS semantic labeling contest \cite{potsdam2012}. It consists of satellite images of the German city Potsdam together with dense label masks for six classes: \texttt{Impervious surfaces}, \texttt{Building}, \texttt{Low vegetation}, \texttt{Tree}, \texttt{Car} and \texttt{Clutter/background}. The orthographic images have a sampling distance of 0.05 m/px. The total dataset consists of 38 tiles with a size of $6000 \times 6000$ px from which we use the same 24 tiles for training as in the original contest. From the remaining tiles, we use 7 for validation and 7 for testing. From the tiles, we randomly crop patches of $256 \times 256$ or $512 \times 512$ pixels. 

\subsection{Encoder and Decoder Implementations}

\begin{figure}[t]
     \centering
     \begin{subfigure}[b]{0.45\textwidth}
         \centering
         \includegraphics[width=\textwidth]{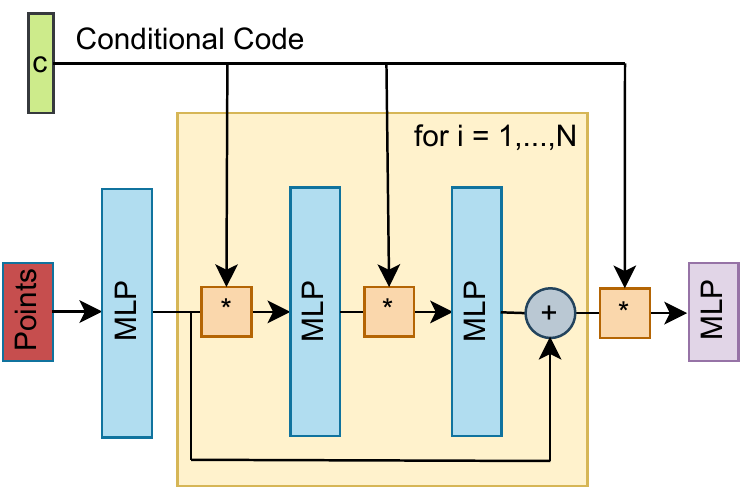}
         \caption{Concat/FiLM Network}
         \label{fig:onet}
     \end{subfigure}
     \hfill
     \begin{subfigure}[b]{0.45\textwidth}
         \centering
         \includegraphics[width=\textwidth]{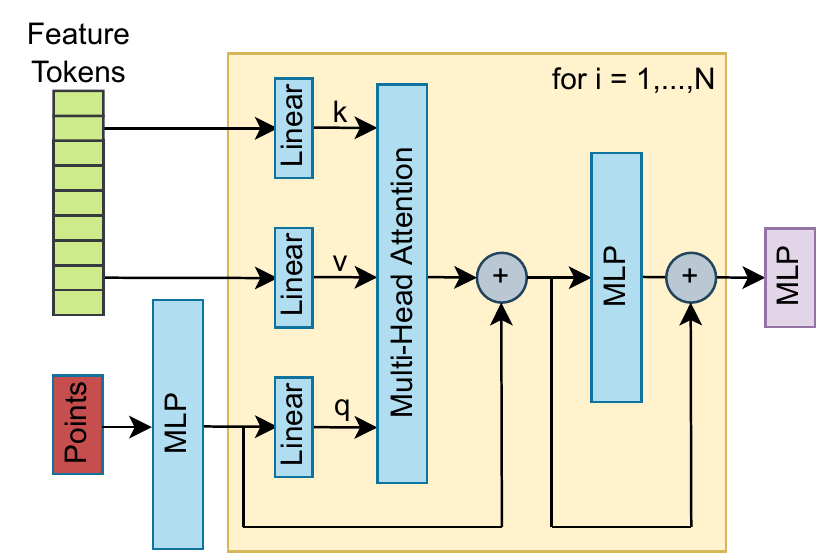}
         \caption{Cross-Attention Transformer}
         \label{fig:transformer}
     \end{subfigure}
     \caption{Our neural network architectures used for the Concat and FiLM conditioning (left) and for the Cross-Attention Transformer (right). The yellow block can be repeated $N$ times. For the Concat approach, the orange block denoted with an asterisk represents a concatination followed by a batchnorm layer. For FiLM, the same block denotes a conditional batchnorm layer. Other batchnorm and layernorm layers have been omitted for clarity.}
     \label{fig3}
\end{figure}

For the Concat and the FiLM decoder, we use a similar neural network architecture, which is based on Occupancy networks \cite{meschederOccupancyNetworksLearning2019} (see Figure \ref{fig:onet}). We use either concatenation plus conventional batchnorm or conditional batchnorm at the designated places in the neural network architecture. For the Cross-Attention conditioning, we use a transformer architecture based on the Scene Representation Transformer \cite{sajjadiSceneRepresentationTransformer2022} (see Figure \ref{fig:transformer}). It uses one multi-head attention module per block. Keys and values are calculated from the feature tokens while the queries are calculated from the point coordinates. We can scale both neural network architectures by repeating the yellow blocks $N$ times or increasing the width of the MLP layers. For all experiments we use a ResNet34 \cite{heDeepResidualLearning2016} backbone as the encoder, pre-trained on ImageNet. Its output feature map has a size of $512 \times 8 \times 8$ for input images with size $256 \times 256 $ pixels and $512 \times 16 \times 16$ for input images with size $512 \times 512 $ pixels respectively. 

\subsection{Points Embedding}

It has been shown that when coordinates are directly used as inputs, neural fields have a bias towards learning low-frequency signals. To counter this, we embed both image coordinates independently into a higher dimensional space by using Fourier features as it is commonly done with neural fields \cite{tancikFourierFeaturesLet2020}:

\begin{equation}
\gamma(x) = (sin(2^0{\pi}x), sin(2^1{\pi}x),...,sin(2^l{\pi}x), cos(2^0{\pi}x), cos(2^1{\pi}x),...,cos(2^l{\pi}x)),
\end{equation}

\noindent where $x$ is an image coordinate and $l$ controls the embedding size.

\subsection{Training Parameters}

The influence of the parameters used in our experiments was evaluated in preliminary runs, based on the validation performance. For all experiments, we choose a fixed learning rate of $1 \times 10^{-4}$ 
for the Adam Optimizer and a batch size of 64. We use horizontal and vertical flipping as data augmentation and perform early stopping based on the IoU metric on the validation set. For all neural field architectures, 512 points are sampled per image and we choose $l=4$ as the size of the points embedding. Empirically, we have found that the results are not sensitive to both these parameters. We have explored scaling the neural field architectures by increasing the number of blocks and the MLP layers' width. With that approach, we use a hidden size of 512 for all MLP layers. One block is used within the Concat and FiLM conditioning network and two blocks are used within the Cross-Attention Transformer. For all architectures, we try to have approximately the same amount of parameters to make a fair comparison. 

\begin{table}[t]
\caption{Results for all examined decoder architectures.}
\begin{tabular}
{  w{c}{2.6cm} | w{c}{2.3cm} | w{c}{1.2cm} | w{c}{1.2cm} | w{c}{1.2cm} | w{c}{1.2cm} | w{c}{1.2cm} }

\hline
  & & \multicolumn{4}{c|}{Image Size} & \\  

\multirowthead{1}{Decoder \\ conditioning} & \multirowthead{1}{Conditional \\ Code Source } & \multicolumn{2}{c}{256} &  \multicolumn{2}{c|}{512} &  \multirowthead{1}{Params}\\

 & & IoU & F-score & IoU & F-score & \\
\hline
Concatenation & global & 0.689 & 0.816 & 0.659 & 0.794 & 2.1M \\ 
			  & local & 0.725 & 0.840 & 0.666 & 0.799 & 2.1M \\ 
			  & global + local & 0.728 & 0.842 & 0.712 & 0.832 & 4.0M\\ \hline
FiLM & global & 0.695 & 0.820 & 0.660 & 0.795 & 2.1M\\ 
			  & local & 0.729 & 0.843 & 0.650 & 0.788 & 2.1M\\ 
			  & global + local & 0.729 & 0.843 & 0.707 & 0.829 & 3.7M\\ \hline
			  
Cross-Attention & feature tokens & 0.758 & 0.862 & 0.754 & 0.860 & 2.6M\\\hline\hline
DeepLabV3+ \cite{chenEncoderDecoderAtrousSeparable2018} & - & 0.760 & 0.863 & 0.763 & 0.866 & 5.4M\\

\end{tabular}
\end{table}

\begin{figure}
\centering
\begingroup
\newcommand{\InclGr}[1]{\frame{\includegraphics[width=2.35cm,height=2.35cm,valign=t]{#1}}}%
\begin{tabular}{c c c c c}
Input Image & Ground Truth & Concat + g & Concat + l & Concat + g/l \\
\InclGr{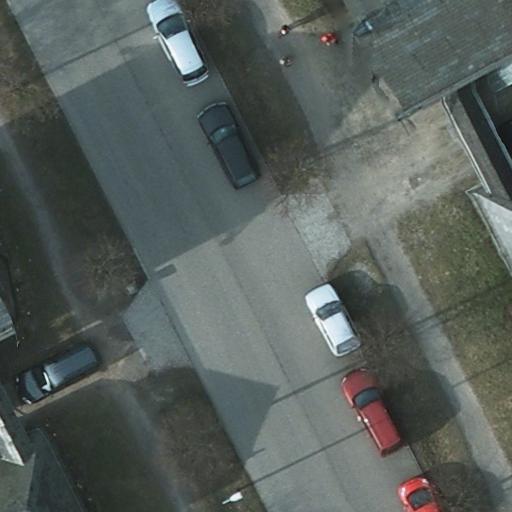} & \InclGr{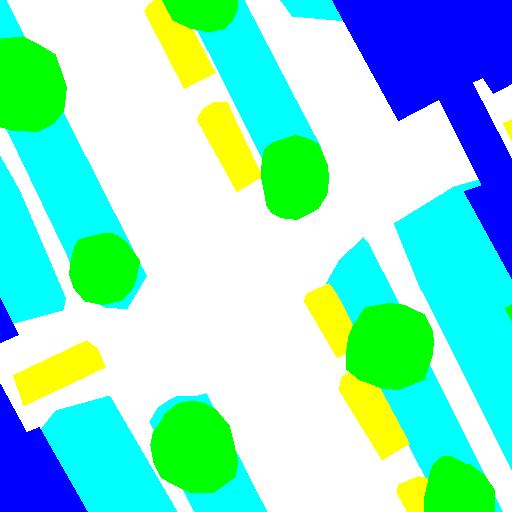} & \InclGr{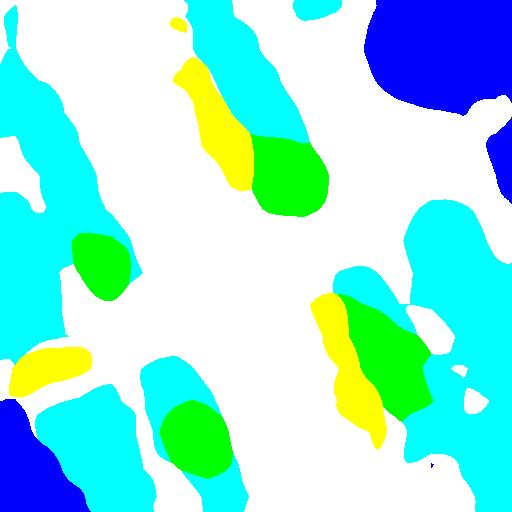} & \InclGr{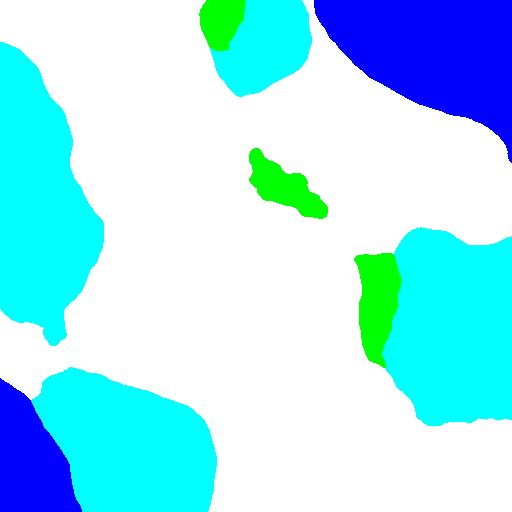} & \InclGr{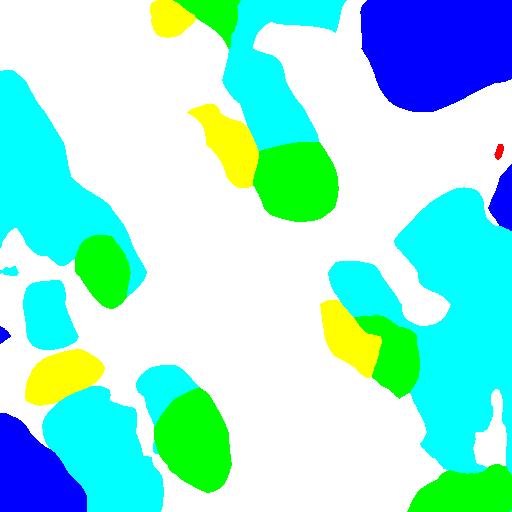} \\
\InclGr{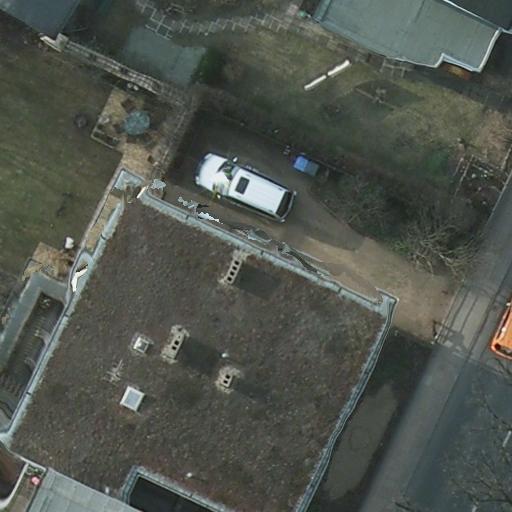} & \InclGr{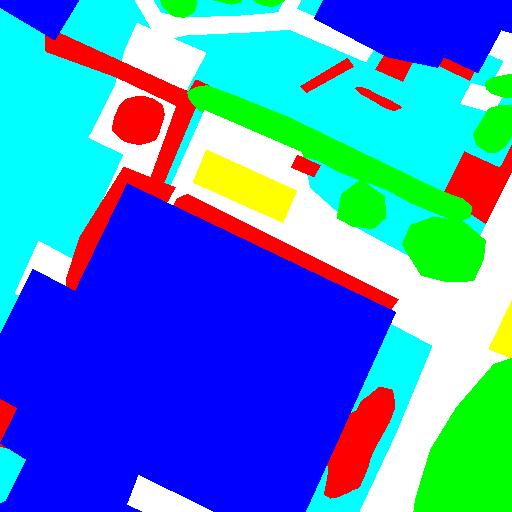} & \InclGr{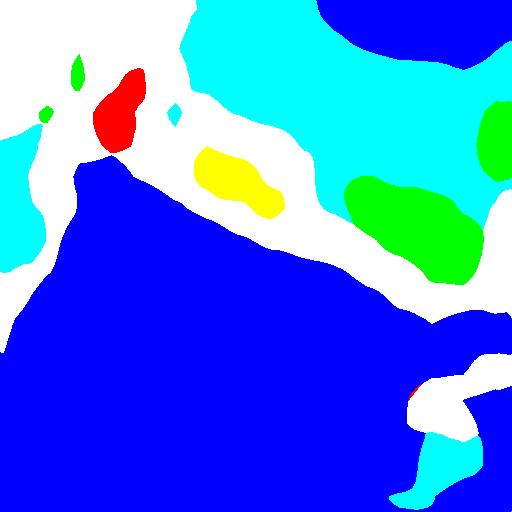} & \InclGr{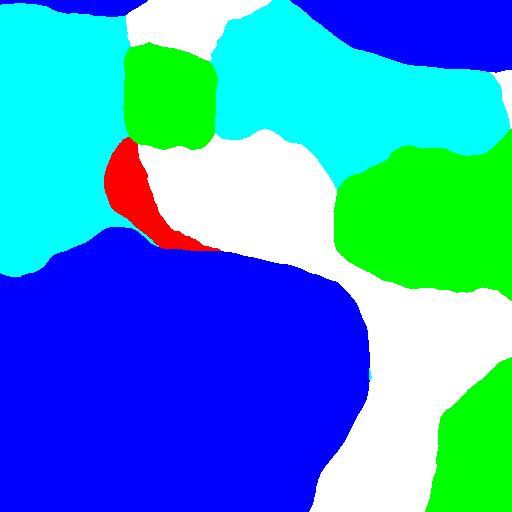} & \InclGr{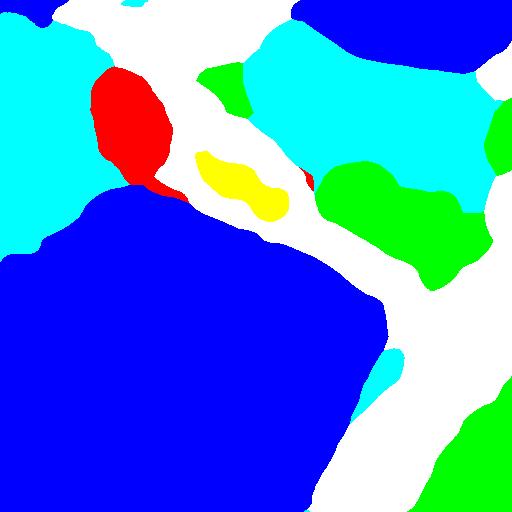} \\
\InclGr{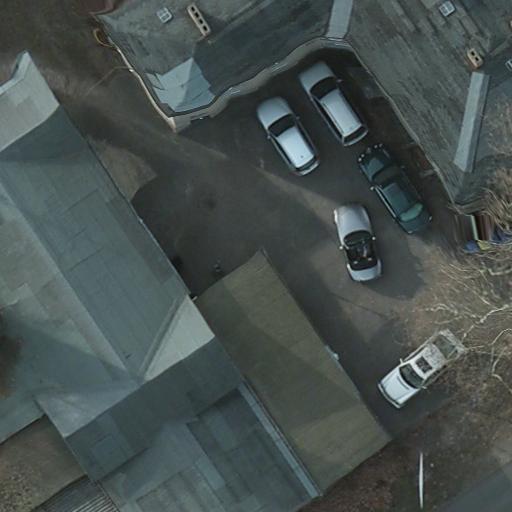} & \InclGr{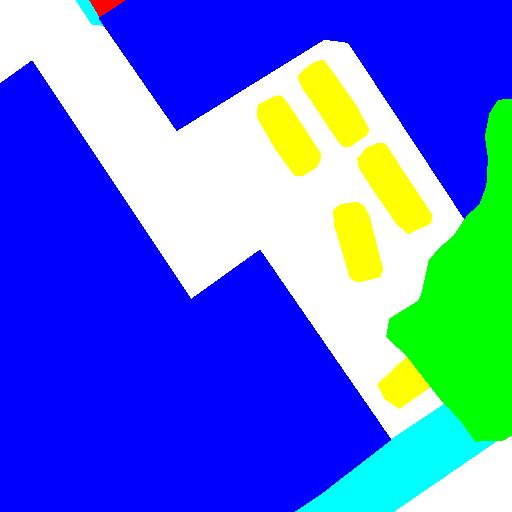} & \InclGr{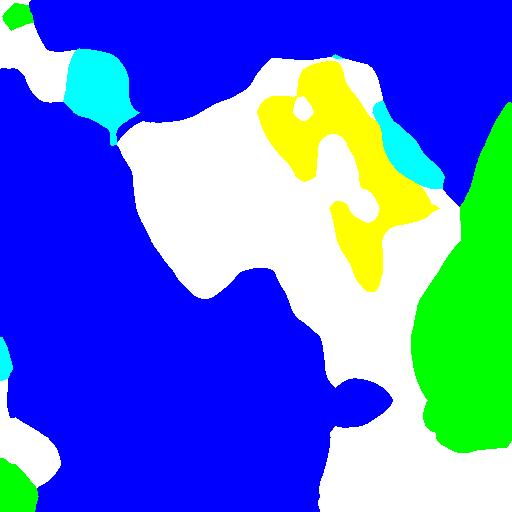} & \InclGr{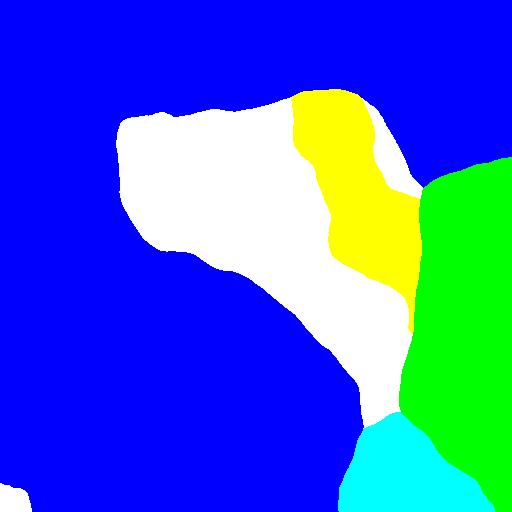} & \InclGr{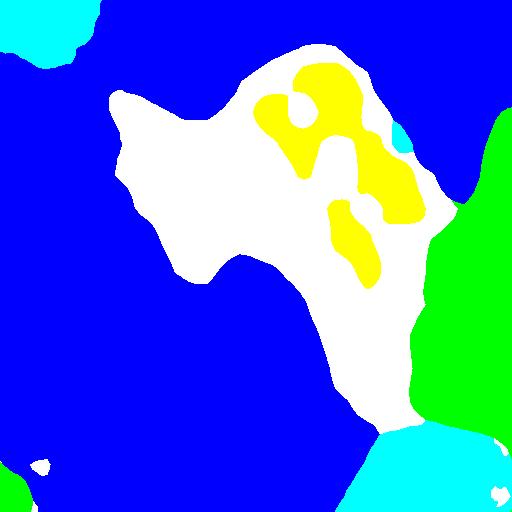} \\
\end{tabular}

\begin{tabular}{c c c c c}
FiLM + g & FiLM + l & FiLM + g/l & Cross-Attn & DeepLabV3+ \\
\InclGr{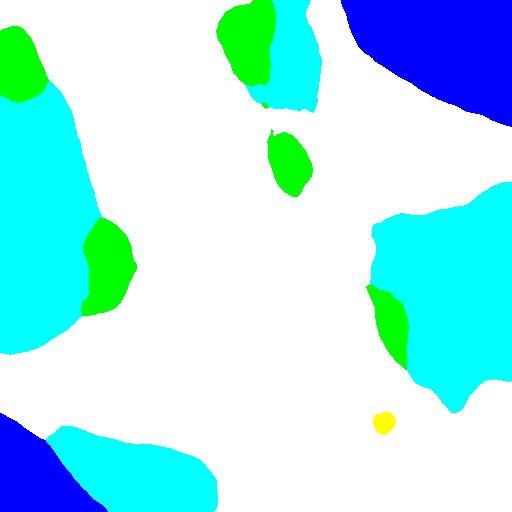} & \InclGr{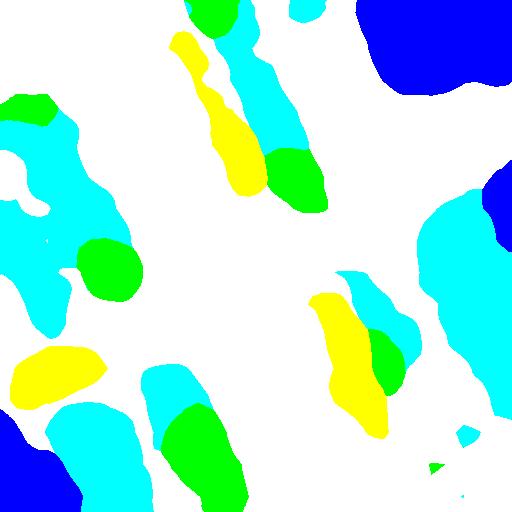} & \InclGr{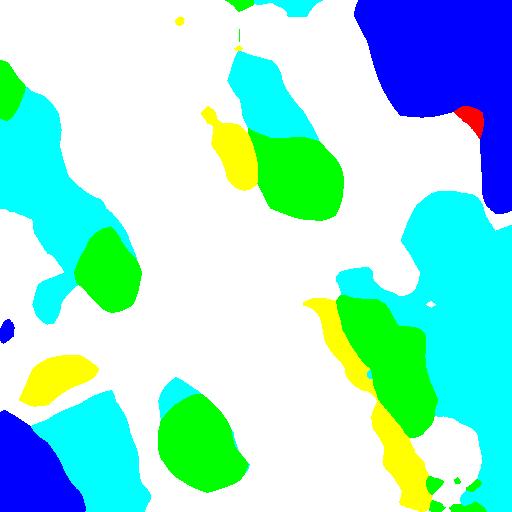} & \InclGr{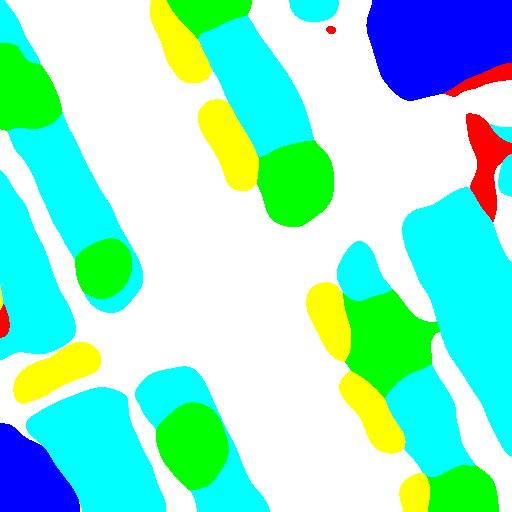} & \InclGr{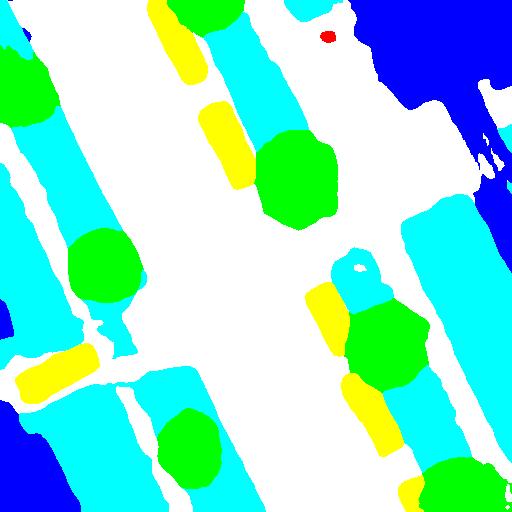} \\
\InclGr{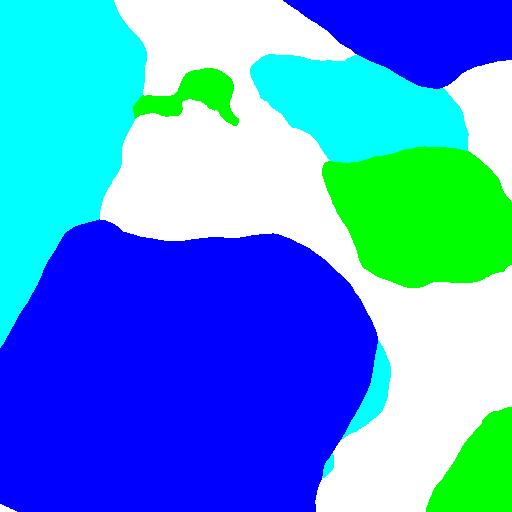} & \InclGr{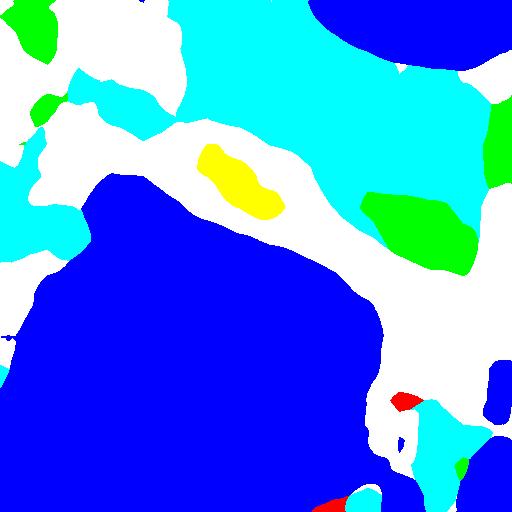} & \InclGr{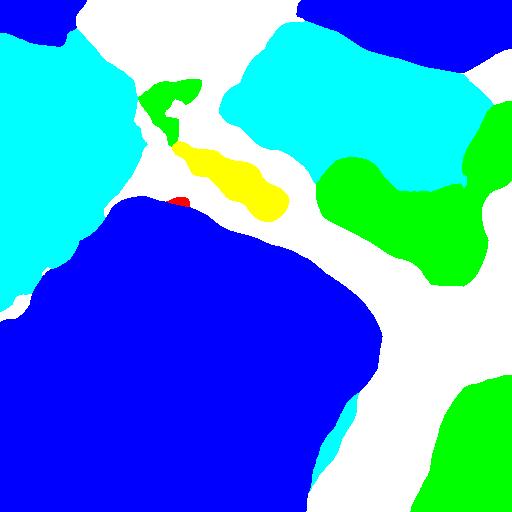} & \InclGr{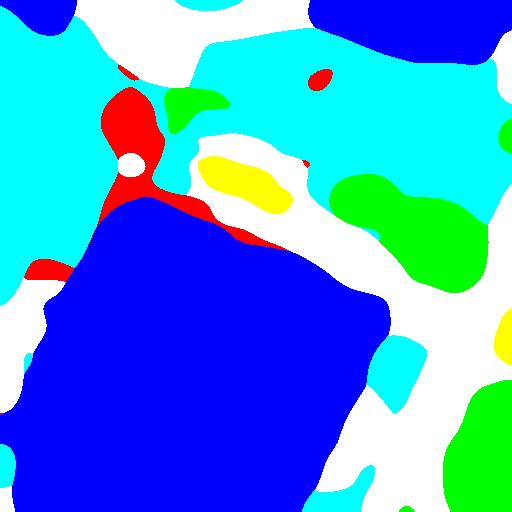} & \InclGr{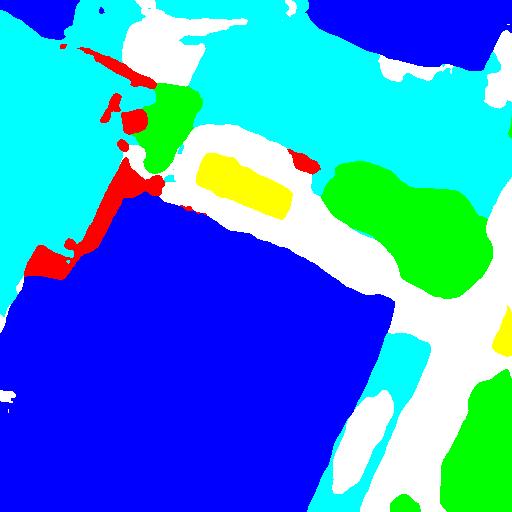} \\
\InclGr{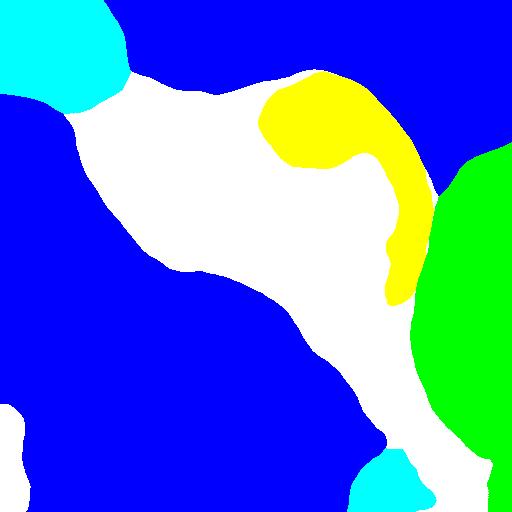} & \InclGr{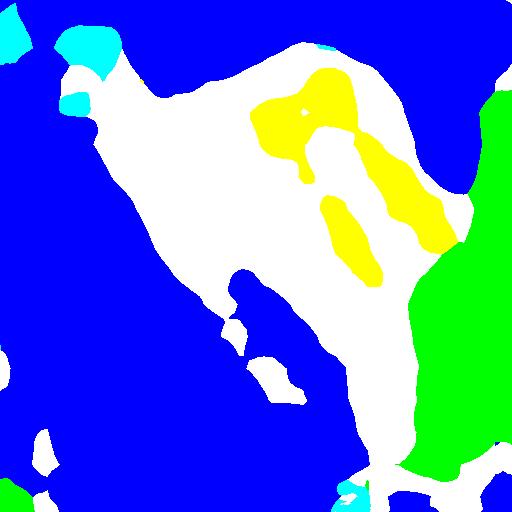} & \InclGr{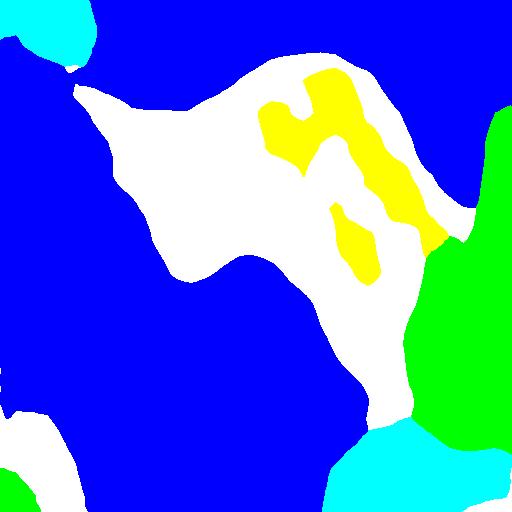} & \InclGr{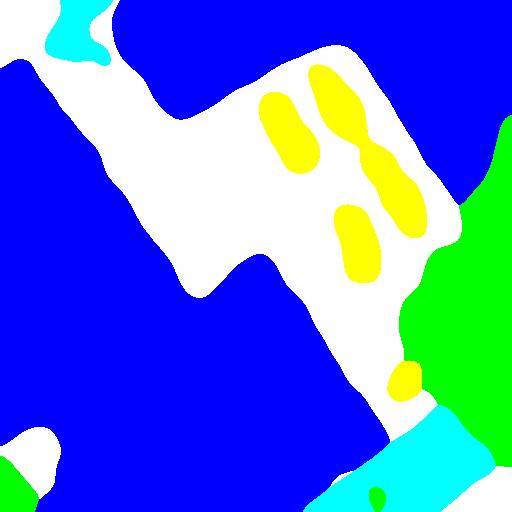} & \InclGr{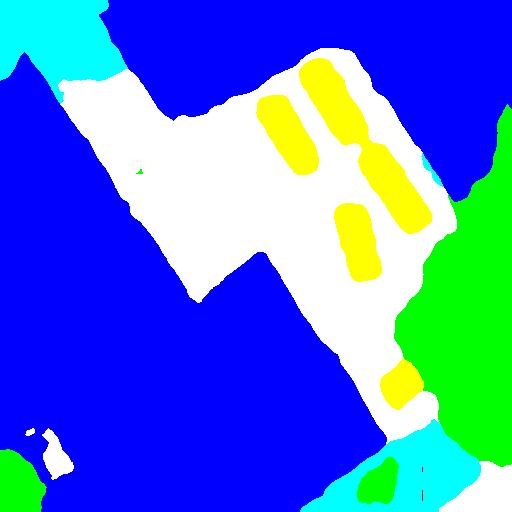} \\
\end{tabular}

\endgroup
\caption{The predictions of all examined decoder architectures on three example images ($512 \times 512 $ px)  from the test set. For Concat and FiLM conditioning, $g$ denotes a global code source, $l$ denotes a local code source and $g/l$ denotes a concatenation of global and local code. The class color code is: white=\texttt{Impervious surfaces}, blue=\texttt{Building}, cyan=\texttt{Low vegetation}, green=\texttt{Tree}, yellow=\texttt{Car}, red=\texttt{Clutter/background}. By comparing the predictions with the ground truth segmentation masks, it can be observed that the ability to represent details, e.g. distinct objects or angular corners, varies greatly between the approaches. Only the Cross-Attention  and the DeepLabV3+ decoders are able to faithfully represent the segmentation masks, while the Concat and FiLm approaches tend to produce overly smooth geometries.}

\end{figure}

\section{Results}

In Table 1 we show the Intersection over Union (IoU), F-Score and the number of parameters for all seven conditioning strategies and two different image sizes on the test set. We also compare our neural field decoder with the DeepLabV3+ \cite{chenEncoderDecoderAtrousSeparable2018} fully convolutional neural network for semantic segmentation which also uses a ResNet34 backbone. In Figure 4 we show the predictions of all decoder architectures for three example images. From the results, we can make multiple key observations.

First, the Concat and FiLM decoders perform very similarly in all aspects, regardless of the conditional code source and the image size.

Second, conditioning via Cross-Attention works best amongst all neural field approaches. Furthermore, it performs similarly to the DeepLabV3+ FCN. Notably, the Cross-Attention decoder has half as much parameters and no access to the intermediate feature maps of the encoder. 

Third, the performance of the Concat and FiLM approaches can be improved by using a combination of global and local features, particularly for larger images. In that case, the performance of both approaches is not much lower compared with the Cross-Attention decoder.

Fourth, the performance of the Concat and FiLM conditioning decreases with larger input images when using global codes. This can be expected, as it is harder to model more geometries in larger images with the same code length. 

Fifth, when using local codes, the performance is also degraded when dealing with larger images. This is unexpected, as the sampling distance (meters per pixel) remains the same and therefore the size of the features should also remain the same. This could be an indication that the individual vectors in the feature map produced by the CNN encoder do not model purely local features, as stated by methods using this approach \cite{chenLearningContinuousImage2021,yuPixelNeRFNeuralRadiance2021}. This is further supported by the fact that modern CNN architectures have very large receptive fields so that one feature vector in the output feature map receives input from the complete input image. In our case, the ResNet34 encoder has a receptive field of 899 pixels which fully covers both our image sizes.

\section{Conclusion}

In this work, we performed a comparative study of neural field conditioning strategies and explored the idea of a neural field-based decoder for 2D semantic segmentation. Our results show that neural fields can have a competitive performance when compared with a classic CNN decoder while requiring even fewer parameters. We also showed that the performance of the neural field is considerably  affected by the conditioning strategy. The best conditioning strategy likely depends on the task. For the task of 2D semantic segmentation, a Cross-Attention-based Transformer is superior to Concat and FiLM conditioning. However, also the combination of local and global conditional codes is a promising approach, as the performance is not much lower. Lastly, for local features, we showed an unexpected degradation in performance when increasing the image size. Further research is required to explain this observation and deduce consequences for local conditioning methods.

%
%
%
\bibliographystyle{splncs04}
\bibliography{bibliography}

\end{document}